\documentclass[10pt, conference]{IEEEtran}
\IEEEoverridecommandlockouts
\usepackage{cite}
\usepackage{amsmath,amssymb,amsfonts}
\usepackage{algorithmic}
\usepackage{graphicx}
\usepackage{textcomp}
\usepackage{xcolor}
\usepackage{booktabs}
\usepackage{tabularx}
\usepackage{multirow} 
\usepackage{subfigure}
\usepackage{subcaption}
\usepackage{adjustbox}
\usepackage{graphicx}
\usepackage{hyperref}
\def\BibTeX{{\rm B\kern-.05em{\sc i\kern-.025em b}\kern-.08em
    T\kern-.1667em\lower.7ex\hbox{E}\kern-.125emX}}
\begin{document}

\title{RE-MCDF: Closed-Loop Multi-Expert LLM Reasoning for Knowledge-Grounded Clinical Diagnosis
}

\author{
	Shaowei Shen\textdagger, Xiaohong Yang\textdagger, Jie Yang, Lianfen Huang$^{*}$, \\ Yongcai Zhang, Yang Zou, Seyyedali Hosseinalipour
	\thanks{The work was supported in part by the National Natural Science Foundation of China under Grant 62371406, the Projects of Fujian Provincial Department of Education JZ250072 and  JAT 251256. Shaowei Shen and Xiaohong Yang are with the School of Informatics, Xiamen University, China. Jie Yang is with the National Institute for Data Science in Health and Medicine, Xiamen University, China. Lianfen Huang is with the Key Laboratory of Intelligent Manufacturing Equipment and Industrial Internet Technology, Fujian Provincial Universities, the School of Information Science and Technology, Xiamen University Tan Kah Kee College, and also with the Department of Informatics and Communication Engineering, Xiamen University, China. Yongcai Zhang is with the School of Medicine, Xiamen University, China. Yang Zou is with the School of Electronic and Information Engineering, Tongji University, China. Seyyedali Hosseinalipour is with the department of Electrical Engineering, University at Buffalo-SUNY, Buffalo, NY, USA (Email: shenshaowei@stu.xmu.edu.cn, xiaohongyang@stu.xmu.edu.cn, leoyang@stu.xmu.edu.cn, lfhuang@xmu.edu.cn, yongcaizhang@stu.xmu.edu.cn, 2152222@tongji.edu.cn and alipour@buffalo.edu).~~ 
	\textdagger~Equal Contribution, $^{*}$ Corresponding Author. }
}

% \par 可以在脚注中换行
\maketitle

\begin{abstract}
Electronic medical records (EMRs), particularly in neurology, are inherently heterogeneous, sparse, and noisy, which poses significant challenges for large language models (LLMs) in clinical diagnosis. In such settings, single-agent systems are vulnerable to self-reinforcing errors, as their predictions lack independent validation and can drift toward spurious conclusions. Although recent multi-agent frameworks attempt to mitigate this issue through collaborative reasoning, their interactions are often shallow and loosely structured, failing to reflect the rigorous, evidence-driven processes used by clinical experts. More fundamentally, existing approaches largely ignore the rich logical dependencies among diseases, such as mutual exclusivity, pathological compatibility, and diagnostic confusion. This limitation prevents them from ruling out clinically implausible hypotheses, even when sufficient evidence is available. To overcome these, we propose RE-MCDF, a relation-enhanced multi-expert clinical diagnosis framework. RE-MCDF introduces a \textit{generation–verification–revision closed-loop} architecture that integrates three complementary components: \textit{(i)} a primary expert that generates candidate diagnoses and supporting evidence, \textit{(ii)} a laboratory expert that dynamically prioritizes heterogeneous clinical indicators, and \textit{(iii)} a multi-relation awareness and evaluation expert group that explicitly enforces inter-disease logical constraints. Guided by a medical knowledge graph (MKG), the first two experts adaptively reweight EMR evidence, while the expert group validates and corrects candidate diagnoses to ensure logical consistency. Extensive experiments on the neurology subset of CMEMR (NEEMRs) and on our curated dataset (XMEMRs) demonstrate that RE-MCDF consistently outperforms state-of-the-art baselines in complex diagnostic scenarios (the source code and datasets are publicly available at \url{https://github.com/shenshaowei/RE-MCDF}).

%Electronic medical records (EMRs), particularly within the neurology subset, are characterized by high heterogeneity, sparsity, and noise, complexities that often impede the diagnostic accuracy of large language models (LLMs). Conventional single-agent approaches are prone to the \textit{echo chamber effect}, where a lack of external validation leads to the self-reinforcement of erroneous conclusions. Meanwhile, existing multi-agent frameworks often rely on superficial interactions that fail to emulate expert disease assessment in real clinical settings. Crucially, the intricate logical relationships between diseases remain under-explored, further hindering diagnostic performance. To address this, we propose RE-MCDF, a relation-enhanced multi-expert clinical diagnosis framework tailored for EMR-based reasoning. We establish a \textit{generation-verification-revision closed-loop mechanism} comprising a primary expert, a laboratory expert, and a multi-relation awareness and evaluation expert group. Guided by a knowledge graph (KG), the two experts dynamically prioritize heterogeneous EMR evidence, while the expert group enforces logical consistency to produce the final diagnosis. Experiments conducted on the neurology subset of the CMEMR dataset (NEEMRs) and our curated dataset (XMEMRs) demonstrate that RE-MCDF significantly outperforms existing baselines in handling complex EMR diagnostic scenarios.
\end{abstract}


\section{Introduction}
The proliferation of large language models (LLMs) has demonstrated immense potential in clinical narrative understanding and generation, particularly for diagnosing cerebrovascular diseases in neurology \cite{ExpertL,Cross-D}. Typically, these models derive diagnostic conclusions by synthesizing multi-source evidence from electronic medical records (EMRs). However, LLMs often lack the complex, multi-step reasoning capabilities required to synthesize heterogeneous and noisy clinical evidence into rigorous diagnoses. Moreover, LLM-driven diagnostic reasoning remains susceptible to hallucinated inferences, incomplete medical coverage, and poorly calibrated confidence, which can lead to unreliable clinical decisions \cite{yang2025towards}. 

To address these, contemporary research has largely followed two major paradigms for improving LLM-based clinical diagnosis: \textit{(i) Prompt-based reasoning strategies}, such as chain-of-thought (CoT), tree of thoughts (ToT), and self-consistency, which aim to bolster reasoning quality and output stability \cite{COT,ToT,Self_C}. However, these reasoning chains are typically expressed in natural language; while they offer some improvement in reliability, they defy systematic verification. Furthermore, in complex scenarios with lengthy records or comorbidities, models remain susceptible to distraction by irrelevant cues, resulting in unreliable candidate generalization and interpretability \cite{Verification,Mind}. \textit{(ii) The integration of knowledge graph (KG) as a structured medical knowledge substrate}, which creates a synergistic LLM-KG paradigm designed to mitigate LLM deficiencies regarding factual coverage and verifiability \cite{gao2025drknows,COG,medikal}. Within this paradigm, prior work generally follows one of two architectural strategies. \textit{(A) Loosely-Coupled Parallel Fusion (LLM$\oplus$KG)}, where the LLM generates candidates or extracts key entities while a parallel process executes retrieval and re-ranking on the KG, with the results from these dual pathways subsequently aggregated at the final stage \cite{mandravickaite2025narrative,MindMap}. 
%For instance, Wei \textit{et al.} proposed a prompting method with KGs to improve LLM reasoning and reduce hallucinations, especially in medical question answering (QA) \cite{MindMap}. Wu \textit{et al}. introduced a framework that injects critical clinical knowledge seeds into LLM prompts to align their reasoning with physician-like decision pathways and improve diagnostic accuracy \cite{ICP}. Jiang \textit{et al.} designed a KG-based framework with LLM-generated hypotheses to enrich query understanding and improve retrieval diversity and accuracy in Chinese medical QA \cite{HyKGE}. In
%addition, Yang \textit{et al.} introduced a medical QA framework that enhances LLMs with KG retrieval and ranking to generate more accurate long-form answers \cite{KGRank}. 
\textit{(B) Deep Collaborative Reasoning (LLM$\otimes$KG)}, which embeds the LLM directly into the graph reasoning process, and entails iterative expansion, node selection, and evidence aggregation for constructing structured reasoning trajectories over the graph \cite{jiang-etal-2025-kg,AdaptBot}. 
%For example, Sun \textit{et al.} proposed a training-free framework where an LLM agent performs beam search over a knowledge graph to enable traceable, correctable, and cost-effective deep reasoning \cite{TOG}. Jin \textit{et al.} designed a framework that enables LLMs to reason iteratively over text-attributed graphs \cite{graph-cot}. Wang \textit{et al.} presented a KG-guided framework that uses iterative symptom inquiry and clinical state tracking for reliable diagnostic reasoning \cite{MedKGI}. 
Beyond such single-agent designs, recent diagnostic systems increasingly explore the synergy between multi-agent collaboration and graph-based reasoning, which has shown promise in handling complex medical decision-making scenarios \cite{Accurate,li-etal-2025-knowledge-aware,10825608}. 

Despite the advancements achieved by the above methods in general tasks, their direct adaptation to the \emph{diagnosis of cerebrovascular diseases in neurology based on EMRs} remains confronted with three fundamental challenges. \noindent \textit{\textbf{Challenge 1:} Static evidence weighting under heterogeneity.} Existing models often adopt a binary \textit{existence-as-match} paradigm for KG evidence, lacking the expert-like ability to dynamically assess the relative importance of heterogeneous and abnormal clinical indicators. As a result, fine-grained evidence prioritization is not achieved. \textit{\textbf{Challenge 2:} Neglect of inter-disease logical constraints.} Most frameworks (single/multi agent) focus on accumulating isolated evidence rather than modeling complex logical relationships (e.g., mutual exclusion and differential ambiguity) \cite{zhou2024high}. This often leads to logically contradictory comorbidities or clinically implausible comorbidity predictions. \textit{\textbf{Challenge 3:} Absence of closed-loop self-correction.} Current unidirectional \textit{input-to-output} workflows lack feedback mechanisms. When faced with ambiguous or conflicting candidates, these systems cannot re-examine original records for differential features, limiting accuracy in complex diagnostic scenarios. 

Motivated by the above challenges, we propose RE-MCDF (Relation-Enhanced Multi-Expert Clinical Diagnosis Framework), entailing the following contributions:
\begin{itemize}
	\item 
	We establish a \textit{generation-verification-revision} mechanism that integrates \textit{primary}, \textit{laboratory}, and \textit{multi-relation group experts}. This iterative loop refines diagnoses through continuous feedback, emulating the rigorous traceability of real-world clinical consultations.
	
	\item 
	To address heterogeneous and noisy EMR, we introduce a \textit{laboratory expert} that dynamically prioritizes abnormal findings. This is coupled with a \textit{KG-driven strategy} to balance diagnostic coverage for long-tail diseases with clinical specificity.
	
	\item
	We devise a multi-perspective evaluation mechanism where experts explicitly model \textit{disease logic integration} and \textit{clinical confusion} relationships. Furthermore, a novel relation-sensitive adjustment strategy with \textit{bounded penalty} is applied, which strictly enforces logical consistency by penalizing conflicting candidates without artificially inflating unsupported diagnoses.
	
	\item 
	Extensive experiments on the cerebrovascular disease subset of CMEMR (NEEMRs) and our own curated diagnostic corpus demonstrate (XMEMRs) that RE-MCDF outperforms state-of-the-art methods, validating the efficacy of our architectural design.
\end{itemize}

\section{Related Work}
\subsection{Disease Prediction from EMRs}
Early EMR-based disease prediction primarily employed structured statistical or machine learning models that leverage tabular data (e.g., demographics, laboratory results, and diagnosis codes) \cite{aminnejad2025predicting,mukherjee2025encounter}. To address data complexity, subsequent work introduced multimodal fusion architectures \cite{11002447} and privacy-preserving federated learning frameworks tailored for heterogeneous, multi-center EMRs \cite{10989538}. More recently, LLMs have demonstrated strong capabilities in extracting insights from unstructured clinical narratives, achieving notable performance in mortality risk stratification \cite{Paging}, automated questionnaire generation \cite{EMRD}, mental health concept classification \cite{cardamone2025classifying}, and adverse drug reaction identification \cite{koon2025unlocking}. Despite these advances, LLM-based approaches often yield \textit{black-box} predictions that lack transparent, evidence-grounded reasoning chains. Moreover, they remain vulnerable to the high heterogeneity, sparsity, and intricate logical relationships among diseases, especially in complex domains like neurology \cite{kang2025llm,lu-naseem-2024-large}. 
Consequently, existing methods struggle to replicate the collaborative, relation-aware reasoning inherent in expert diagnosis. To bridge this gap, we propose a \textit{relation-enhanced multi-expert framework} that explicitly models inter-disease constraints and \textit{enforces verifiable, closed-loop reasoning}.

\subsection{KG-Enhanced Reasoning with LLMs}
To enhance factual grounding and mitigate model hallucination, recent efforts integrate KGs into LLM-based reasoning. Early single-agent approaches inject external knowledge via KG-augmented prompting \cite{MindMap}, in-context knowledge seeds \cite{ICP}, retrieval-augmented generation \cite{HyKGE,KGRank}, or iterative graph traversal \cite{TOG,graph-cot}, and \cite{MedKGI} further incorporates KG-constrained differential diagnosis with information-guided inquiry. While these methods improve accuracy, they place the full burden of evidence synthesis, logical inference, and error correction on a single agent, leading to error accumulation and limited reasoning depth. To distribute cognitive load, multi-agent frameworks have been proposed, where specialized agents collaborate on diagnosis \cite{MedLA}. For instance, \cite{Magic} uses multi-agent debate to refine graph weights, while \cite{KERAP} employs linkage, retrieval, and prediction agents for zero-shot diagnosis. Some even leverage multi-LLM consensus for KG construction and validation \cite{Clinical}. However, most systems treat diseases as atomic labels rather than nodes in a relational network, focusing on surface-level agreement while neglecting disease-specific logical constraints (e.g., mutual exclusion). RE-MCDF addresses this by embedding relation-aware logic into a closed-loop framework, ensuring diagnostic conclusions are both knowledge-grounded and logically consistent.

\section{Proposed Method}
\subsection{Problem Formulation}
We consider a set of EMRs, each represented as $\mathbf{x} = \{x_{\text{cc}}, x_{\text{hpi}}, x_{\text{pmh}}, x_{\text{pe}}, x_{\text{ae}}\}$, which comprises the \textit{chief complaint}, \textit{history of present illness}, \textit{past medical history}, \textit{physical examination}, and \textit{auxiliary examination}. 
Inspired by the workflow of neurological expert consultations, we construct a multi-expert diagnostic network composed of a \textit{primary expert} ($\mathcal{A}_{\text{pri}}$), a \textit{laboratory expert} ($\mathcal{A}_{\text{lab}}$), and a \textit{multi-relation awareness and evaluation group} ($\mathcal{A}_{\text{multi}}$). Additionally, we leverage a medical knowledge graph (MKG) $\mathcal{G} = (\mathcal{V}, \mathcal{M})$ as a structured knowledge substrate, where $\mathcal{V}$ and $\mathcal{M}$ denote the set of nodes (e.g., diseases, symptoms, and examinations) and the semantic relations connecting them. The objective is to generate a Top-$k$ diagnostic set $\mathcal{D}_k = \{d_1, \dots, d_k\}$. To this end, as illustrated in Fig. 1, RE-MCDF establishes a \textit{closed-loop diagnostic process}:
\begin{equation}
	\mathcal{D}_k = \mathcal{F}_\text{fusion} \left( \mathcal{A}_\text{pri}, \mathcal{A}_\text{lab}, \mathcal{A}_\text{multi} \mid \mathbf{x}, \mathcal{G} \right) \xrightarrow{\text{feedback}} \mathcal{A}_\text{pri},
	\label{eq:pipeline}
\end{equation}
where $\mathcal{A}_{\text{pri}}$ generates initial diagnostic hypotheses and evidence pairs, $\mathcal{A}_{\text{lab}}$ identifies abnormal indicators and computes dynamic weights for these entities, which are mapped to corresponding nodes in $\mathcal{V}$, and $\mathcal{A}_{\text{multi}}$ analyzes inter-disease relationships to determine the final diagnosis. Furthermore, in the event of diagnostic conflicts, an evaluative feedback mechanism is triggered, applying logical penalties to the initial hypotheses of $\mathcal{A}_{\text{pri}}$ for post-hoc confidence calibration. This relation-aware calibration helps ensure that diagnostic decisions are supported by traceable clinical evidence and explicit logical justification.
\begin{figure*}[t]
	\centering
	\includegraphics[width=0.8\textwidth, height=\textheight, keepaspectratio]{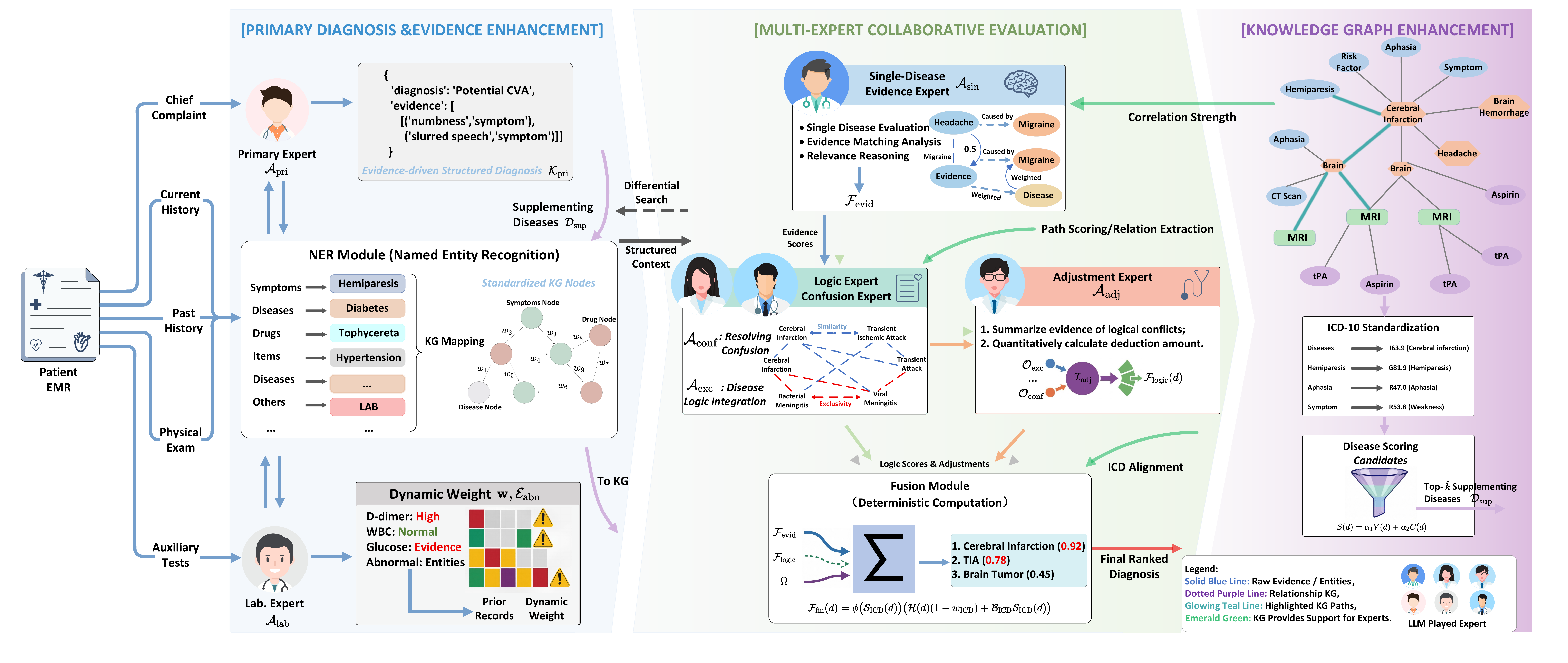}
	\caption{Architecture of RE-MCDF. It consists of a primary expert that generates initial diagnosis–evidence pairs from EMRs, and a laboratory expert that dynamically weights heterogeneous clinical indicators (e.g., abnormal laboratory findings). A MKG is then used to expand candidate coverage and provide structured relational paths to four collaborative experts: single-disease assessment, disease logic integration, confusion detection, and logical adjustment. Together, these experts enforce logical consistency through relation-aware scoring and trigger closed-loop feedback when conflicts are detected (e.g., mutually exclusive diseases with similar confidence), enabling traceable and clinically plausible diagnostic reasoning.}
	\label{fig:model}
\end{figure*}

\subsection{Primary Diagnosis and Evidence Enhancement}
﻿
In clinical practice, physicians infer diagnoses by synthesizing latent and heterogeneous evidence from the patient's chief complaint, laboratory results, and medical history. However, the scenarios we confront are characterized by high evidence heterogeneity, which renders traditional schemes such as named entity recognition (NER) and relation extraction (RE) suboptimal. Consequently, we introduce $\mathcal{A}_{\text{pri}}$ to facilitate the generation of traceable evidence chains for subsequent validation through structured reasoning. Specifically, $\mathcal{A}_{\text{pri}}$ executes a constrained CoT reasoning process: given a summarized medical record $\mathbf{s} = \text{Summarize}(\mathbf{x})$, the model outputs structured diagnosis-evidence pairs:
$
	\mathcal{K}_{\text{pri}} = \{ (d_i, \mathcal{E}_i) \}_{i=1}^{N}, \mathcal{E}_i = \{e_{i1}, \dots, e_{im}\}
	\label{eq:primary_output},
$
where $d_i$ denotes a candidate disease and $\mathcal{E}_i$ represents its supporting evidence entity. Let $\mathcal{E}=\{\mathcal{E}_1,...,\mathcal{E}_N\}$. To resolve semantic ambiguity, $\mathcal{E}$ are aligned to a standard medical comparison table $\mathcal{U}$, resulting in a standardized entity set $\mathcal{E}_{\text{std}} = \left\{ e \mid \text{type}(e) \in \mathcal{T} \right\}$, where $\mathcal{T}$ is a predefined set of functional categories, ensuring consistent semantic grounding and enables structured downstream reasoning.

Furthermore, in this scenario, the potential pathological evidence and its corresponding diagnostic weight vary substantially across individual cases; thus, a \textit{one-size-fits-all} weighting scheme, prevalent in existing literature, often leads to erroneous judgments by neglecting case-specific critical indicators. To address this, we integrated NER with $\mathcal{A}_{\text{lab}}$ to collaborate with downstream experts, which identifies abnormal indicators and assigns prioritized weights to pathological findings, thereby amplifying the influence of salient clinical evidence on downstream reasoning. Specifically, it first performs structural processing of the EMR via a medical NER model \cite{wang-etal-2021-improving} to extract medical entities. Then, based on the patient's basic information summary $\mathbf{s}_{\text{base}}$ and summary of inspection results $\mathbf{s}_{\text{exam}}$, $\mathcal{A}_{\text{lab}}$ perform the following calculations:
\begin{equation}
	\begin{cases}
		\mathbf{w} = \{w_{\text{type}(e)}\}_{\text{type}(e) \in \mathcal{T}} = \text{Norm}\left( \psi_{\text{wei}}(\mathbf{s}_{\text{base}}, \mathbf{s}_{\text{exam}}, \mathcal{E}_\text{std}) \right), \\
		\mathcal{E}_{\text{abn}} = \psi_{\text{abn}}(\mathbf{s}_{\text{base}}, \mathbf{s}_{\text{exam}}, \mathcal{E}_\text{std}),
	\end{cases}
\end{equation}
where, for each entity $e \in \mathcal{E}$, $w_{\text{type}(e)}$ denotes its base weight, $\text{Norm}(\cdot)$ is a normalization operation, which ensures $\sum_{\text{type}(e)\in \mathcal{T}}w_{\text{type}(e)}=1$, while $\psi_{\text{wei}}$ and $\psi_{\text{abn}}$ represent the LLM-implemented reasoning logic of $\mathcal{A}_{\text{lab}}$ for weight generation $\mathbf{w}$ and abnormality identification $\mathcal{E}_{\text{abn}}$, respectively. 

Although LLM-based approaches utilizing existing information can cover a majority of diagnostic scenarios, intrinsic knowledge limitations, particularly in smaller-scale models, inevitably lead to the omission of diagnoses, resulting in insufficient coverage of long-tail diseases \cite{Dynamic,SurveyA,contrastive}. To mitigate this deficiency, we leverage the established efficacy of MKG to provide an external knowledge base that compensates for the LLM's inherent shortcomings. Therefore, in our methodology, we integrate a MKG to supplement diagnoses that may be overlooked by the LLM through structured knowledge. First, for $\mathcal{E}_{\text{std}}$, the entity-disease association strength is defined as:
\begin{equation}
	\Gamma(e, d) = \begin{cases}
		1, & \text{if } \text{dist}_{\mathcal{G}}(e,d)\leq 3, \\
		0, & \text{otherwise},
	\end{cases}
	\label{eq:edge_connection}
\end{equation}
where $\text{dist}_\mathcal{G}(e,d)$ denotes the shortest path length between $e$ and $d$ in the MKG. Subsequently, we map the entity weights to prioritize abnormal evidence:
\begin{equation}
	w'_{e} = \begin{cases}
		1.5w_{\text{type}(e)}, & e \in \mathcal{E}_{\text{abn}}, \\
		w_{\text{type}(e)}, & \text{otherwise}.
	\end{cases}
	\label{eq:entity_weight_mapping}
\end{equation}
Then, we propose a disease scoring function that balances evidence coverage, specificity, and connectivity:
\begin{equation}
	S(d) = \alpha_1 V(d) + \alpha_2 C(d),\hspace{2mm}\alpha_1 + \alpha_2 = 1,
	\label{eq:disease_scoring}
\end{equation}
where $V(d) =
	\sum\limits_{e \in \mathcal{E}_{\text{con}}(d, n)} 
	 \dfrac{w'_e}{\text{dist}_\mathcal{G}(e,d) + \varepsilon} /
	\sum\limits_{e \in \mathcal{E}_{\text{con}}(d, n)} w'_e
$ represents the weighted connectivity strength derived from the shortest path, $\mathcal{E}_{\text{con}}(d) = \left\{ e \in \mathcal{E}_{\text{std}} \mid 1 \leq \text{dist}_\mathcal{G}(e,d) \leq 3 \right\}$, and $\varepsilon=10^{-8}$ is a smoothing factor. Also, the weighted evidence coverage is defined as $C(d) = 
\sum\limits_{e \in \mathcal{E}_{\text{std}}} w'_e \Gamma(e, d)/
\sum\limits_{e \in \mathcal{E}_{\text{std}}} w'_e
$. Finally, the Top-$\hat{k}_{\text{sup}}$ diseases ranked by $S(d)$ are selected to form the supplementary candidate set $\mathcal{D}_{\text{sup}}$.

\subsection{Multi-Expert Collaborative Evaluation}
As discussed above, although initial disease–evidence pairs can be generated by a single agent, such models struggle to reason over complex inter-disease dependencies, including mutual exclusivity, comorbidity versus causality, and diagnostic confusion. To overcome these limitations, recent work has explored collaborative reasoning paradigms that combine multiple independent perspectives with external evidence constraints \cite{MAMM,muma,SciAgents}. Building upon this, we introduce $\mathcal{A}_{\text{multi}}$, consisting of four specialized agents: \textit{single-disease assessment expert} ($\mathcal{A}_{\text{sin}}$), \textit{disease logic integration expert} ($\mathcal{A}_{\text{exc}}$), \textit{confusion expert} ($\mathcal{A}_{\text{conf}}$), and \textit{adjustment expert} ($\mathcal{A}_{\text{adj}}$).

\subsubsection{Single-Disease Evidence Expert} 
As the foundational stage of the multi-expert collaborative evaluation, $\mathcal{A}_{\text{sin}}$ assesses the plausibility of the initial diagnostic evidence and the alignment between candidate diseases and the critical indicators identified by $\mathcal{A}_{\text{lab}}$. Specifically, $\mathcal{A}_{\text{sin}}$ utilizes structural grounding from $\mathcal{G}$ to quantify the degree of match between a disease and the patient's clinical presentation, which processes three core categories of input: \textit{(i)} the set of candidate diseases $\mathcal{D}_\text{cand}$ derived from $\mathcal{K}_\text{pri}$ and $\mathcal{D}_\text{sup}$, \textit{(ii)} $\mathcal{E}_{\text{abn}}$ identified by $\mathcal{A}_{\text{lab}}$, and
\textit{(iii)} the disease-evidence connectivity paths derived from $\mathcal{G}$. 
The evidence consistency score is defined as:
\begin{equation}
	\mathcal{F}_{\text{evid}}(d) = \mathcal{A}_{\text{sin}}\left(\mathcal{C}_{\mathcal{G}}(d), \mathcal{E}_{\text{abn}}, \mathbf{s}\right),
\end{equation}
where $\mathcal{A}_{\text{sin}}(\cdot)$ is a output based on existing input of $\mathcal{A}_{\text{sin}}$, and $\mathcal{C}_{\mathcal{G}}(d) = \mathcal{N}_{\text{dir}}(d) \cup \mathcal{N}_{\text{path}}(d, \mathcal{E}_\text{std})$ provides the structured knowledge context for disease $d$. Here, $\mathcal{N}_{\text{dir}}(d) = \{(d, r_i, t_i) \in \mathcal{G}\}$ represents the set of triples directly incident to $d$, where $r_i$ denotes the medical relation (e.g., \textit{has\_symptom} or \textit{belongs\_to}) and $t_i$ is the corresponding target entity (e.g., a specific symptom or disease category). Meanwhile, $\mathcal{N}_{\text{path}}(d, \mathcal{E}_\text{std})$ encompasses all triples situated along the shortest paths from evidence entities $e_j \in \mathcal{E}_\text{std}$ to the candidate disease $d$. The scoring process explicitly emphasizes two factors:
\begin{itemize}
	\item \textit{Path Semantic Quality}: A weight-decay mechanism is applied based on path directness: direct relations (e.g., ``Hypertension $\xrightarrow{\text{presentation}}$ High Blood Pressure'') receive higher weights, while indirect multi-hop paths (e.g., ``Chronic Hypertension $\xrightarrow{\text{causes}}$ Cerebral Arteriolar Hyalinosis $\xrightarrow{\text{results in}}$ Deep White Matter Lacunar Infarction'') are progressively attenuated.
	
	\item \textit{Priority of Abnormal Entities}:  $\mathcal{E}_{\text{abn}}$ is introduced (see \eqref{eq:entity_weight_mapping}), ensuring that clinically critical findings exert greater influence on the final score.
	
\end{itemize}

\subsubsection{Multi-Disease Relation Awareness} 

To better capture inter-disease dynamics (e.g., mutual exclusion, coexistence, causality, and differential ambiguity), we design $\mathcal{A}_{\text{exc}}$ and $\mathcal{A}_{\text{conf}}$, which reason over relational structures extracted from $\mathcal{G}$, based on two specialized relationship sets: the \textit{pathological compatibility relationship} $\mathcal{R}_{\text{exc}}$ and the \textit{clinical confusion relationship} $\mathcal{R}_{\text{conf}}$. Notably, they produce a \textit{logical confidence score} $\mathcal{F}_{\text{logic}}(d)$ of candidate diseases, which serves as the primary metric for finalizing the diagnostic set $\mathcal{D}_k$. 

To operationalize the logical constraints within $\mathcal{A}_{\text{exc}}$, we align the clinical diagnostic process with the fundamental principles of \textit{collective exhaustivity} and \textit{mutual exclusivity}. Specifically, guided by the taxonomic hierarchy of \cite{zhou2024high}, we focus on the \textit{pathological compatibility} within the initial candidate set. Unlike generalized semantic similarity, this relationship characterizes the \textit{biological compatibility} between entities, a critical mechanism for pruning logically implausible disease combinations. These constraints are explicitly derived from the \textit{pathological subtype edges} in $\mathcal{G}$, formalized as:
$
	\mathcal{R}_{\text{exc}} = \big\{ (d_i, d_j, e) \mid \hat{r}(d_i, d_j) \in \mathcal{G}, \ q \in \mathcal{P}_{\text{str}} \big\},
	\label{eq:exclusive_relations}
$
where $\hat{r}$ denotes predefined pathological typing relations,
$\mathcal{P}_{\text{str}}$ denotes the set of structured evidence paths, and $q$ is a path indicating pathological exclusivity (e.g., ``HER2 Positive $\leftarrow$[Type]$\rightarrow$ HER2 Negative''). This formulation captures three types of strict exclusivity: \textit{(i)} Mutually exclusive molecular subtypes of the same disease (e.g., EGFR-mutant vs. ALK-rearranged); \textit{(ii)} Mutually exclusive pathological classification standards (e.g., Type 1 vs. Type 2 Diabetes); \textit{(iii)} Chronologically distinct disease stages (e.g., Acute vs. Old Myocardial Infarction). Accordingly, $\mathcal{A}_{\text{exc}}$ reasons on the structured input $\mathcal{C}_{\text{exc}}$ based on the predefined output conditions $\mathcal{P}_{\text{exc}}$, yielding $\mathcal{O}_{\text{exc}}$:
\begin{equation}
	\begin{cases}
		\mathcal{C}_{\text{exc}} &= \bigcup\limits_{(d_i,d_j) \in \mathcal{R}_{\text{exc}}} \big\{ \text{evid}_{\text{path}}(d_i,d_j) \big\} \cup \big\{ \mathcal{C}_{\text{pt}}, \mathcal{D}_{\text{cand}} \big\}, \\
		\mathcal{O}_{\text{exc}} &= \mathcal{A}_{\text{exc}}\left(\mathcal{C}_{\text{exc}}, \mathcal{P}_{\text{exc}}\right),
	\end{cases}
\end{equation}
where $\mathcal{C}_{\text{pt}}$ represents the patient context,  $\text{evid}_{\text{path}}(d_i,d_j)$ denotes the MKG path substantiating exclusivity between $d_i$ and $d_j$, and $\mathcal{A}_{\text{exc}}(\cdot)$ is the output of $\mathcal{A}_{\text{exc}}$. While, the implementation of clinical confusion relations is more challenging, as relying solely on MKG is insufficient for comprehensive coverage. Therefore, we design a dual confusion discrimination mechanism for $\mathcal{A}_{\text{conf}}$: \textit{(i)} Explicit differential diagnosis edges $\overline{r}$ in MKG; \textit{(ii)} Attribute overlap analysis based on core clinical features (symptoms/imaging/biochemistry), defined as:
$
	\mathcal{R}_{\text{conf}} = \big\{ (d_i, d_j) \mid \overline{r}(d_i, d_j) \in \mathcal{G} \lor |\xi(d_i, d_j)| \geq 2 \big\},
	\label{eq:confusion_relations}
$
where $\xi(d_i, d_j) = \Phi(d_i) \cap \Phi(d_j)$ represents the \textit{feature overlap set}, characterizing the diagnostic intersection between two candidate diseases, and $\Phi(d)$ denotes the set of core clinical features associated with disease $d$. Consequently, the input evidence structure $\mathcal{C}_{\text{conf}}$ and the verified output $\mathcal{O}_{\text{conf}}$ of $\mathcal{A}_{\text{conf}}$ are defined as:
	\begin{equation}
	\hspace{-4.5mm}
	\resizebox{0.45\textwidth}{!}{$
	\begin{cases}
	\mathcal{C}_{\text{conf}} &= \bigcup\limits_{(d_i,d_j) \in \mathcal{R}_{\text{conf}}} \left\{
	\begin{aligned}
		&\mathcal{I}_{\text{share}} = \xi_{\text{type}(e)}(d_i,d_j), \\
		&\mathcal{I}_{\text{diff}} = \Phi(d_i) \triangle \Phi(d_j), \\
		&\mathcal{I}_{\mathcal{G}} = \text{path}_{\mathcal{G}}(d_i,d_j),
	\end{aligned}
	\right\} \cup \mathcal{C}_{\text{pt}}, \\
	\mathcal{O}_{\text{conf}} &= \mathcal{A}_{\text{conf}}\left(\mathcal{C}_{\text{conf}}, \mathcal{P}_{\text{conf}}\right),
\end{cases}
		$}\hspace{-3.15mm}
\end{equation}
where $\mathcal{A}_{\text{conf}}(\cdot)$ is the output of $\mathcal{A}_{\text{conf}}$, $\mathcal{I}_{\text{share}}$ represents the evidence derived from shared features, $\mathcal{I}_{\text{diff}}$ denotes the \textit{discriminative attributes} calculated via the symmetric difference $\triangle$,  $\mathcal{I}_{\mathcal{G}}$ denotes explicit associative paths in $\mathcal{G}$, and the expert assessment protocol $\mathcal{P}_{\text{conf}}$ enforces a rigorous \textit{logical verification} mechanism. Then, the relational sets $\mathcal{O}_{\text{exc}}$ and $\mathcal{O}_{\text{conf}}$ are aggregated into a comprehensive context $\mathcal{I}_{\text{adj}}$, which serves as the input set for $\mathcal{A}_{\text{adj}}$:
\begin{equation}
	\mathcal{I}_{\text{adj}} = \mathcal{D}_{\text{cand}} \cup \mathcal{O}_{\text{exc}} \cup \mathcal{O}_{\text{conf}} \cup \mathcal{C}_{\text{pt}}.
\end{equation}

To ensure clinical robustness, $\mathcal{A}_{\text{adj}}$ processes this context to generate the logical confidence score $\mathcal{F}_{\text{logic}}(d)$, subject to three strict properties:
\begin{itemize}
	\item \textit{Bounded Penalty:} $\mathcal{F}_{\text{logic}}(d) \in [0, 1]$, preventing evidence-free inflation.
	
	\item \textit{Relation-Sensitive Attenuation:} 
	\textit{(i)} For $\mathcal{O}_{\text{exc}}$, weaker candidates incur stronger penalties ($\mathcal{F}_{\text{logic}} \to 0$). 
	\textit{(ii)} For $\mathcal{O}_{\text{conf}}$, all relevant candidates are synchronously attenuated based on clinical feature uncertainty.
	
	\item \textit{Default Consistency:} For candidates not involved in verification relations, the system maintains original confidence by setting $\mathcal{F}_{\text{logic}}(d) = 1$.
\end{itemize}

Finally, guided by the aforementioned constraints and inputs, $\mathcal{A}_{\text{adj}}$ execute clinical logical reasoning, with the output strictly constrained to a \textit{piecewise structure}:
\begin{equation}
	\mathcal{F}_{\text{logic}}(d) = \begin{cases} 
		\mathcal{A}_{\text{adj}}(d, \mathcal{I}_{\text{adj}}), & \text{if } d \in \text{Dom}(\mathcal{O}_{\text{exc}} \cup \mathcal{O}_{\text{conf}}), \\
		1, & \text{otherwise},
	\end{cases}
	\label{mud}
\end{equation}
where $\text{Dom}(\cdot)$ denotes the \textit{domain of influence} (i.e., the set of disease entities involved in identified logical relationships), and  $\mathcal{A}_{\text{adj}}(\cdot)$ is the output of $\mathcal{A}_{\text{adj}}$.

\subsubsection{Final diagnosis generation}
We implement a \textit{multi-tiered fusion strategy} to compute the finalized scores for candidate diseases. We first calculates a \textit{core composite score} $\mathcal{H}(d) = \beta_1\mathcal{F}_{\text{evid}}(d) + \beta_2\mathcal{F}_{\text{logic}}(d) + \beta_3\Omega(d)$, where $\Omega(d)$ denotes the graph connectivity score between the disease and evidence entities, $\sum_{i=1}^3 \beta_i = 1$, and then incorporates an \textit{ICD-10 standardization penalty mechanism}\cite{zhou2024high} to generate a diagnostic ranking that aligns with clinical practice:
\begin{equation}
	\begin{aligned}
		\mathcal{F}_{\text{fin}}(d) = \phi\big(\mathcal{S}_{\text{ICD}}(d)\big)\big(\mathcal{H}(d)(1-w_{\text{ICD}}) + \mathcal{B}_{\text{ICD}} \mathcal{S}_{\text{ICD}}(d)\big),
	\end{aligned}
	\label{eq:final_fusion}
\end{equation}
where $\mathcal{S}_{\text{ICD}}(d)$ represents the normalized ICD-10 standardization similarity score, $\mathcal{B}_{\text{ICD}}$ serves as the weight for ICD-related features and $w_{\text{ICD}}$ controls the contribution of the ICD alignment term. Also, the \textit{penalty function} $\phi(\cdot)$ is defined as:
\begin{equation}
	\phi\big(\mathcal{S}_{\text{ICD}}(d)\big) = 
	\begin{cases} 
		1, & \text{if } \mathcal{S}_{\text{ICD}}(d) \geq \tau_{\text{thr}}, \\
		\left(\frac{\mathcal{S}_{\text{ICD}}(d)}{\tau_{\text{thr}}}\right)^\gamma, & \text{otherwise},
	\end{cases}
	\label{eq:penalty_function}
\end{equation}
where $\tau_{\text{thr}}$ is the predefined ICD compliance threshold and $\gamma$ denotes the penalty intensity coefficient.  Finally, diseases are ranked by 
$\mathcal{F}_{\text{fin}}$, and the Top-$k$ are returned as $\mathcal{D}_k$.

%Finally, after multi-expert collaborative evaluation, we employ deterministic weighted aggregation to output candidate diagnosis disease scores and select the Top-$k$ diseases as the final diagnosis, defined as:
%\begin{equation}
%	\mathcal{F}_{\text{fin}}(d) = \mu_1\mathcal{F}_{\text{evid}} + \mu_2\mathcal{F}_{\text{logic}} + \mu_3\Omega + \mu_4\mathcal{S}_{\text{ICD}},
%	\label{eq:final_fusion}
%\end{equation}
%where the weighting coefficients satisfy $\sum_{i=1}^4 \mu_i = 1$, and are optimized via grid search to ensure the dominance of clinical evidence.

\section{Experiments and Evaluations}
In this section, we first detail the datasets and experimental settings. Subsequently, we present a comparative analysis between RE-MCDF and representative methods. Finally, we conduct ablation studies to verify the effectiveness of the specific modules within RE-MCDF.
\begin{table}[t]
	\centering
	\footnotesize
	\caption{Statistical comparison of datasets. Average input length aggregates all structured clinical fields (e.g., history, examinations, and results).}
	\label{tab:shuom}
	\begin{tabular}{lccc}
		\toprule
		\textbf{Dataset} & \textbf{Samples} & \textbf{Avg. Labels} & \textbf{Avg. Len.} \\
		\midrule
		NEEMRs (Public) & 1,001 & 2.75 & 582.4 \\
		XMEMRs (Self-constructed)            & 1,024 & 5.48 & 2,779.2 \\
		\bottomrule
	\end{tabular}
	\vspace{-3mm}
\end{table}

\subsection{Datasets}
	To evaluate the performance of RE-MCDF across both standardized benchmarks and  real-world clinical scenarios, we conducted experiments on two distinct datasets. 

\begin{table*}[t]
	\centering
	% \footnotesize
	\caption{Experimental results on NEEMRs and XMEMRs datasets across different LLM backbones. All metrics are reported in percentage (\%), where \textbf{R}, \textbf{P} and \textbf{F1} stands for Recall, Precision, and F1 Score.}
	\label{tab:main_results}
	\renewcommand{\arraystretch}{0.8}
	\begin{adjustbox}{max width=\textwidth}
	\setlength{\tabcolsep}{2pt}
	
	\newcolumntype{Y}{>{\centering\arraybackslash}X}
	
	\begin{tabularx}{\textwidth}{l l *{3}{Y} *{3}{Y} | *{3}{Y} *{3}{Y}}
		\toprule
		\multicolumn{2}{c}{\multirow{4.5}{*}{\textbf{Methods}}}& \multicolumn{6}{c|}{\textbf{Qwen2.5-7B-Instruct}} & \multicolumn{6}{c}{\textbf{GLM-4-9B}} \\
		\cmidrule(lr){3-8} \cmidrule(lr){9-14}
		& & \multicolumn{3}{c}{\textbf{NEEMRs}} & \multicolumn{3}{c|}{\textbf{XMEMRs}} & \multicolumn{3}{c}{\textbf{NEEMRs}} & \multicolumn{3}{c}{\textbf{XMEMRs}} \\
		\cmidrule(lr){3-5} \cmidrule(lr){6-8} \cmidrule(lr){9-11} \cmidrule(lr){12-14}
		& & R & P & F1 & R & P & F1 & R & P & F1 & R & P & F1 \\ 
		\midrule
		
		\multirow{3}{*}{\textit{LLM-only}} 
		& CoT \cite{COT} & 37.80 & 39.03 & 38.40 & 35.01 & 32.85 & 33.90 & 41.29 & 39.50 & 40.37 & 37.33 & 35.80 & 36.55 \\
		& ToT \cite{ToT} & 42.23 & 38.68 & 40.38 & 33.71 & 31.01 & 32.30 & 39.94 & 36.71 & 38.26 & 38.82 & 35.62 & 37.15 \\
		& Sc-CoT \cite{Self_C} & 44.23 & 30.44 & 36.06 & 31.03 & 34.62 & 32.73 & 39.58 & 35.19 & 37.25 & 38.10 & 35.00 & 36.48 \\ 
		\midrule
		
		\multirow{3}{*}{\textit{LLM$\oplus$KG}} 
		& MindMap \cite{MindMap} & 45.54 & 41.91 & 43.65 & 38.46 & 36.45 & 37.43 & 43.54 &
		40.22 & 41.82 & 35.48 & 38.40 & 36.88 \\
		& ICP \cite{ICP} & 40.96 & 37.87 & 39.36 & 34.01 & 32.50 & 33.24 & 37.25 & 36.78 & 37.01 & 38.10 & 34.86 & 36.41 \\
		& KG-Rank \cite{KGRank} & 38.34 & 35.22 & 36.71 & 35.39 & 32.42 & 33.83 & 40.00
		& 33.33 & 36.36 & 34.19 & 37.38 & 35.72\\
		\midrule
		
		\multirow{2}{*}{\textit{LLM$\otimes$KG}} 
		& MedIKAL \cite{medikal} & 41.32 & 41.63 & 41.47 & 38.24 & 38.32 & 38.28 & 40.71 & 39.45 & 40.07 & 38.77 & 37.91 & 38.34 \\
		& Graph-CoT \cite{graph-cot} & 45.43 & 41.66 & 43.46 & 40.64 & 37.06 & 38.76 & 42.07 & 38.78 & 40.36 & 38.45 & 37.87 & 38.16\\ 
		\midrule
		
		\multirow{2}{*}{\textit{Multi-Agent}} 
		& MAGIC \cite{Magic} & 35.61 & 33.32 & 34.42 & 34.88 & 33.26 & 34.05 & 36.67 & 36.67 & 36.67 & 38.30 & 35.17 & 36.67 \\
		& KERAP \cite{KERAP} & 33.87 & 40.32 & 36.81 & 33.14 & 37.10 & 35.01 & 36.25 & 38.78 & 37.47 & 37.88 & 36.70 & 37.28 \\
		\midrule
		
		\multicolumn{2}{l}{\textbf{RE-MCDF (Ours)}} & \textbf{46.13} & \textbf{42.28} & \textbf{44.11} & \textbf{42.33} & \textbf{38.78} & \textbf{40.48} & \textbf{44.20} & \textbf{40.47} & \textbf{42.25} & \textbf{42.23} & \textbf{38.69} &
		\textbf{40.38} \\
		\bottomrule
	\end{tabularx}
	\end{adjustbox}
	\vspace{-3mm}
\end{table*}

\noindent\textbf{NEEMRs (Public Benchmark):}
We utilize the neurology subset of the CMEMR \cite{medikal}, where each record contains core clinical fields including the chief complaint, basic information, history of present illness (HPI), past medical history (PMH), physical examination, and auxiliary examination.

\noindent\textbf{XMEMRs (Self-constructed Clinical Dataset):}
Due to the scarcity of high-quality neurology EMR datasets in existing research, we constructed XMEMRs, a neurology-focused EMR dataset designed to reflect high-complexity real-world clinical scenarios. The dataset was collected from the Department of Neurology at a Grade-A Tertiary Hospital between 2024 and 2025\footnote{The First Affiliated Hospital of Xiamen University.}. All records underwent rigorous quality control by three senior neurologists, who excluded cases with missing information, inconsistent documentation, or insufficient diagnostic evidence. The remaining records were then standardized and manually annotated with confirmed diagnoses. Compared to NEEMRs, XMEMRs contains more comprehensive primary examination data, including sequential laboratory results, imaging reports, and electrocardiograms (ECGs), making it more challenging and clinically realistic. A statistical comparison between the two datasets is provided in Table~\ref{tab:shuom}.

\subsection{Implementation Details}
\noindent \textbf{Medical Knowledge Graph:} We utilize CPubMed-KGv2 (1.7 million nodes, 3.9 million relations). Shortest path calculations are accelerated via Neo4j GDS. For ICD-10 standardization, we follow \cite{medikal,Fan2024AIHI}, adopting a fuzzy string matching algorithm with a similarity threshold set to 0.5.

\noindent \textbf{Model Configuration:} Qwen2.5-7B-Instruct and GLM-4-9B (bfloat16 precision) serves as the backbone model, loaded via vLLM with the temperature set to 0 to ensure deterministic outputs. For NER, we employ the medical-domain RaNER model \cite{wang-etal-2021-improving} to identify 9 categories of entities (i.e., $\mathcal{T}=\{\text{sym}, \text{dis}, \text{dru}, \text{bod}, \text{ite}, \text{equ}, \text{mic}, \text{dep}, \text{pro}\}$), and we use Bge-small-zh for embeddings.

\noindent \textbf{Baseline}: We compared RE-MCDF with four series of baseline methods: LLM-only, LLM$\oplus$KG, LLM$\otimes$KG and multi-agent combination KG. \textit{\textbf{LLM-only}}: They only use the LLMs' internal knowledge for reasoning, including CoT \cite{COT}, ToT \cite{ToT} and  Sc\_Cot \cite{Self_C}. \textit{\textbf{LLM$\oplus$KG}}: We selected there representative works, namely MindMap \cite{MindMap}, ICP \cite{ICP} and KG-Rank \cite{KGRank}, all of which are aimed at medical question-answering and reasoning tasks. \textbf{\textit{LLM$\otimes$KG}}: We select MedIKAL \cite{medikal} and Graph-Cot \cite{graph-cot}. \textbf{\textit{Multi-Agent combination KG}}: We implemented a multi-agent framework for the medical field combined with KG, namely MAGIC \cite{Magic}, KERAP \cite{KERAP}.

\subsection{Performance Comparison with Baseline Methods}

Table~\ref{tab:main_results} compares RE-MCDF against the baselines across four technical categories. Overall, our framework consistently achieves state-of-the-art (SOTA) performance across all backbones. Compared to LLM-only approaches, RE-MCDF exhibits a substantial leap; on NEEMRs (Qwen2.5), it achieves an F1 of 44.11\%, surpassing Sc-CoT and ToT by gains of 8.05\% and 3.73\%. This highlights the limitations of vanilla LLMs, where clinical noise often impedes reasoning without structured validation. Furthermore, RE-MCDF consistently outperforms knowledge-enhanced (e.g., MindMap), tightly-coupled (e.g., MedIKAL), and multi-agent (e.g., KERAP) systems. By explicitly modeling logical interdependencies, RE-MCDF effectively overcomes the knowledge noise and conflicting reasoning paths prevalent in these advanced architectures, ensuring more precise evidence alignment in complex diagnostic tasks. The superiority of RE-MCDF remains consistent when scaling to the GLM-4 backbone. In the more challenging XMEMRs scenario, RE-MCDF (F1 = 40.38\%) significantly outperforms the strongest knowledge-enhanced baselines, such as MedIKAL (38.34\%) and Graph-CoT (38.16\%). Notably, while most baselines exhibit unstable cross-dataset performance, suffering significant drops on XMEMRs under Qwen2.5 or inconsistent fluctuations under GLM-4, RE-MCDF consistently achieves the highest F1 and superior robustness. On NEEMRs and XMEMRs, outperforming all competitors, thereby confirming its resilience in complex real-world diagnostics.

\begin{table}[htbp]
	\centering
	% \footnotesize
	\caption{Ablation study results of the RE-MCDF on NEEMRs and XMEMRs (\textit{ Qwen2.5; all metrics in \%}).}
	\label{tab:ablation}
	\setlength{\tabcolsep}{6pt}
	\renewcommand{\arraystretch}{0.8}
	\begin{adjustbox}{max width=\columnwidth}
	\begin{tabular}{l|ccc|ccc}
		\toprule
		\multirow{2.5}{*}{\textbf{Config.}} 
		& \multicolumn{3}{c|}{\textbf{NEEMRs}} 
		& \multicolumn{3}{c}{\textbf{XMEMRs}} \\
		\cmidrule(lr){2-4} \cmidrule(lr){5-7}
		& \textbf{R} & \textbf{P} & \textbf{F1} 
		& \textbf{R} & \textbf{P} & \textbf{F1} \\
		\midrule
		\textbf{RE-MCDF} & \textbf{46.13} & \textbf{42.28} & \textbf{44.11} & \textbf{42.33} & \textbf{38.78} & \textbf{40.48} \\
		\midrule
		w/o $\mathcal{G}_{\text{MKG}}$ & 44.12 & 40.43 & 42.20 & 40.14 & 36.70 & 38.34 \\
		w/o $\mathcal{A}_{\text{lab}}$ & 43.50 & 40.46 & 41.70 & 39.49 & 36.11 & 37.73 \\
		w/o $\mathcal{A}_{\text{sin}}$ & 44.13 & 40.45 & 42.21 & 40.18 & 36.73 & 38.37 \\
		w/o $\mathcal{A}_{\text{rel}}$ & 45.72 & 41.90 & 43.72 & 40.27 & 36.94 & 38.54 \\
		\midrule
		LLM-direct & 41.79 & 40.78 & 41.28 & 36.96 & 33.89 & 35.35 \\
		\bottomrule
	\end{tabular}
	\end{adjustbox}
\end{table}

\vspace{-3mm}
\subsection{Ablation Study-I}
As illustrated in Table~\ref{tab:ablation}, the full RE-MCDF model consistently achieves the optimal performance across all datasets (including methods for directly predicting diseases using LLM). Interestingly, $\mathcal{A}_{\text{lab}}$ emerges as the most critical component for diagnostic accuracy; its removal leads to the most significant F1 drop of 2.41\% and 2.75\% on NEEMRs and XMEMRs. This highlights the importance of dynamic evidence weighting in clinical practice. The $\mathcal{G}_{\text{MKG}}$ and $\mathcal{A}_{\text{sin}}$ modules also play vital roles, particularly in ensuring diagnostic coverage for complex cases, with their absence resulting in F1 declines of 2.14\% and 2.11\% for XMEMRs, respectively. Furthermore, $\mathcal{A}_{\text{rel}}$ (i.e., $\mathcal{A}_{\text{exc}} + \mathcal{A}_{\text{conf}}$) module ensures logical consistency by filtering out contradictory diagnoses (e.g., mutually exclusive stroke types). Disabling this module causes the F1 to fall to 38.54\% on XMEMRs. Manual review confirms that $\mathcal{A}_{\text{rel}}$ successfully resolved logical conflicts in 17.8\% of analyzed cases, preventing common diagnostic pitfalls. These results demonstrate that the synergy between structured knowledge, expert-based evidence weighting, and logical relationship mining is essential for superior clinical decision support.
\begin{figure}[t]
	\centering
	\vspace{-3mm}
	\subfigure[]{
		\includegraphics[trim=0.1cm 0.05cm 0.1cm 0cm, clip, width=0.46\columnwidth]{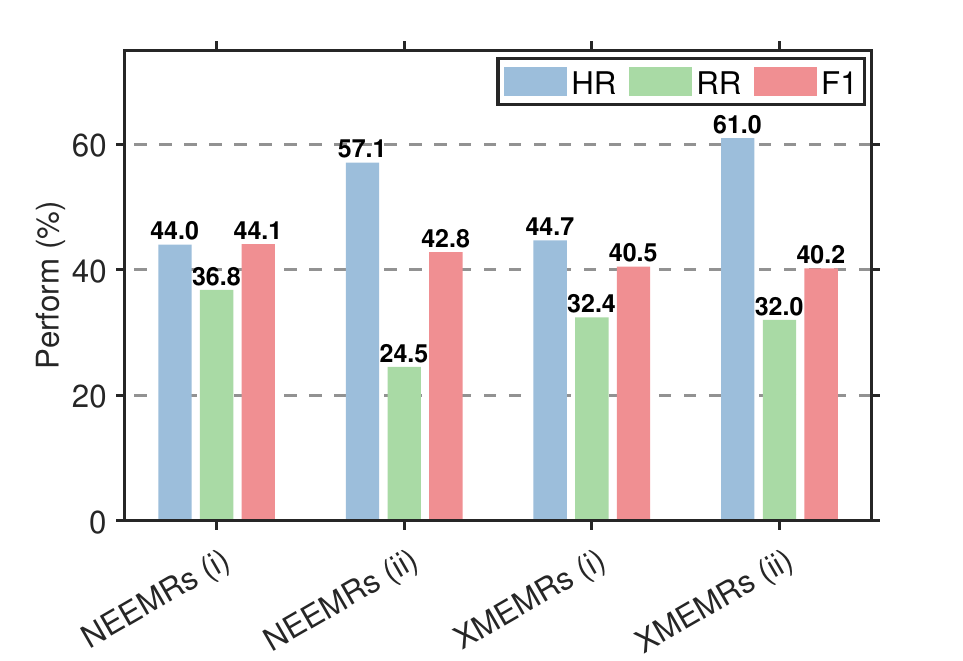}
	}
	\subfigure[]{
		\includegraphics[trim=0.1cm 0.05cm 0.1cm 0cm, clip, width=0.46\columnwidth]{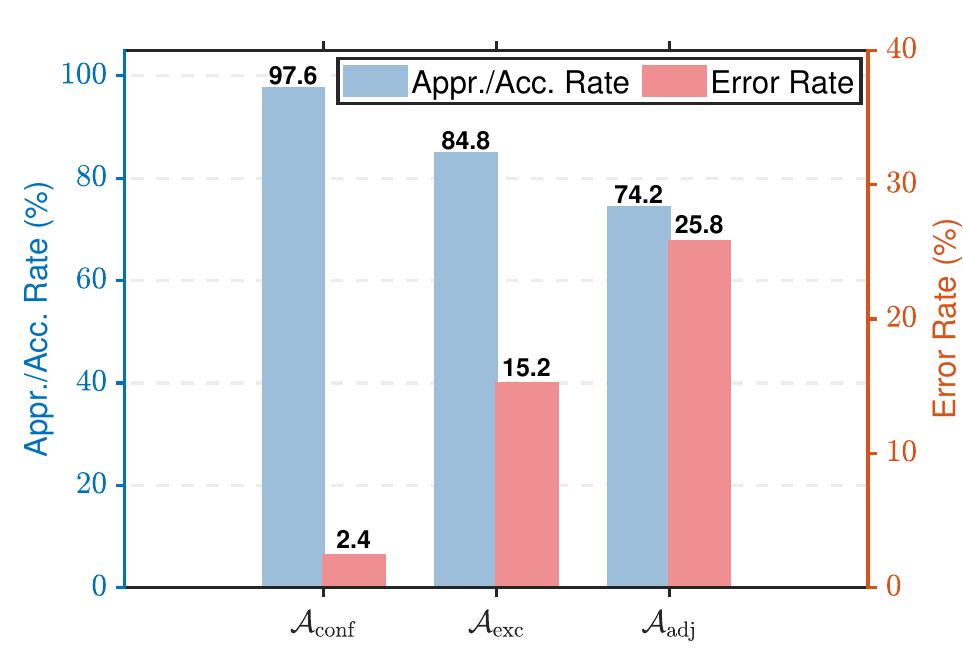}
	}
	\vspace{-3mm}
	\caption{Performance analysis and human evaluation. Left subplot: Impact of MKG supplement depth $\hat{k}_\text{sup}$ on retrieval and verification ($\hat{k}_\text{sup}$ = 1 for (i), $\hat{k}_\text{sup}$ = 2 for (ii)). Right subplot: Manual evaluation of reasoning trajectories, where $\mathcal{A}_{\text{conf}}$ and  $\mathcal{A}_{\text{exc}}$ are measured by physician approval rate, and $\mathcal{A}_{\text{adj}}$ by accuracy rate (\textit{based on Qwen2.5}).}
	\label{fig:2}
	\vspace{-2.5mm}
\end{figure}

\subsection{Ablation Study-II}
To evaluate the effectiveness of the MKG supplement mechanism, we define two metrics: hit rate (HR), which measures initial coverage (i.e., whether one supplement matches ground truth via fuzzy matching), and retained-and-hit rate (RR), which measures the proportion of supplemented candidates that are successfully verified by $\mathcal{A}_{\text{sin}}$ and  $\mathcal{A}_{\text{rel}}$ and retained in the final Top-$\hat{k}$. As shown in Fig.~\ref{fig:2} (a), while HR increases monotonically with $\hat{k}_{\text{sup}}$, RR stagnates or even declines significantly. For the NEEMRs dataset, increasing $\hat{k}_{\text{sup}}$ from 1 to 2 leads to a sharp 12.3\% absolute drop in RR (36.8\% $\to$ 24.5\%). This phenomenon reveals an ``attention dilution" effect: a single high-confidence supplement ($\hat{k}_{\text{sup}}$ = 1) allows $\mathcal{A}_{\text{sin}}$, $\mathcal{A}_{\text{rel}}$ to focus on precise evidence matching and logical verification. Conversely, higher $\hat{k}_{\text{sup}}$ values introduce irrelevant candidates that act as noise, overwhelming the experts with complex relational paths and increasing the risk of false suppression of true diagnoses. This degradation is more pronounced on the NEEMRs dataset due to its lower diagnostic density (2.75 labels/case vs. 5.48 in XMEMRs). In such sparse scenarios, extra candidates are more likely to lack clinical relevance, misallocating the multi-expert system's computational focus and hindering the retention of correct diagnoses. Across both datasets, $\hat{k}_{\text{sup}}$ = 1 yields the optimal F1, validating our \textit{precision over breadth} design principle. This aligns with clinical heuristics where experienced physicians prioritize 1–2 key differential diagnoses for in-depth verification.
\begin{figure}[t]
	\centering
	\vspace{-1.5mm}
	\includegraphics[width=3.5in]{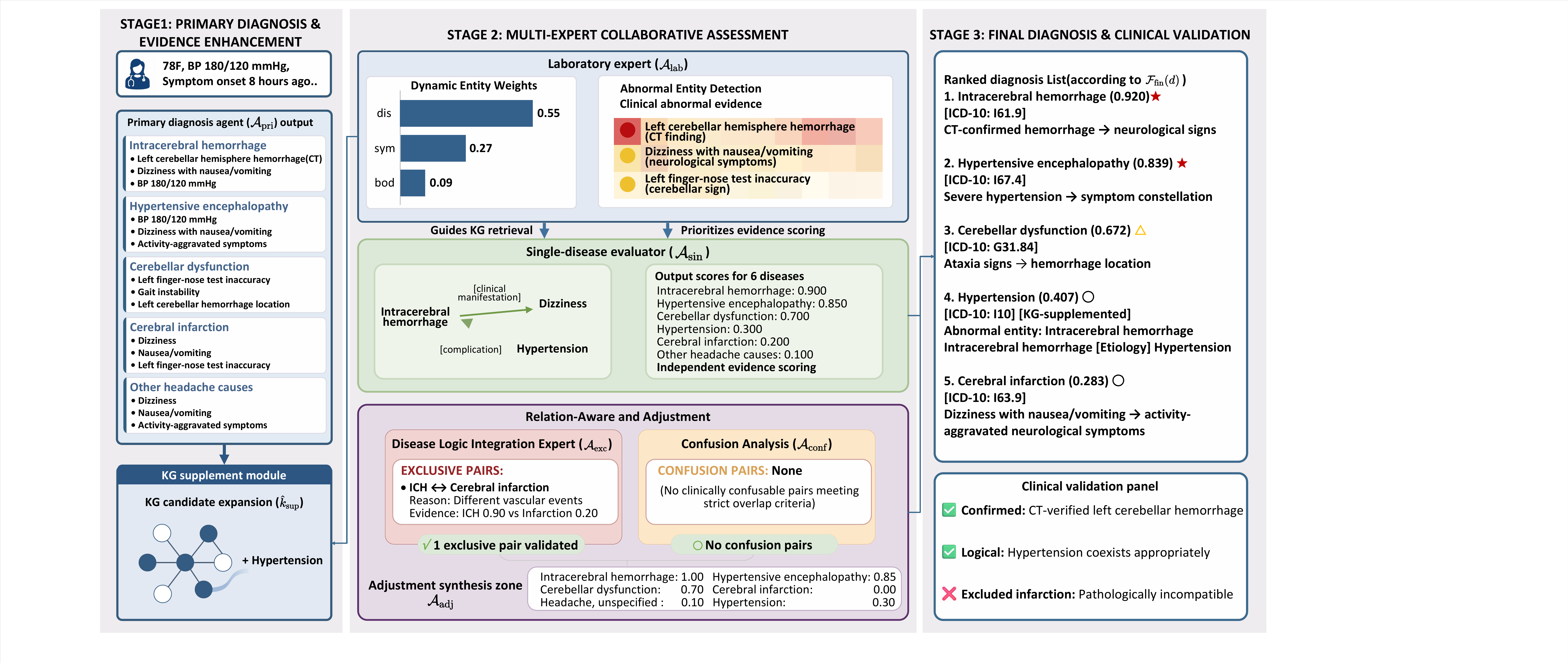}
	\caption{Illustrative case study of RE-MCDF: Integrating clinical evidence with relation-aware reasoning.}
	\label{fig:example}
	\vspace{-3mm}
\end{figure}

\vspace{-1.2mm}
\subsection{Ablation Study-III}
\subsubsection{Fine grained analysis of expert collaboration mechanism} To verify clinical reliability, senior neurologists conducted a blind review of reasoning trajectories sampled from XMEMRs, focusing on logical consistency and weighting rationality. As shown in Fig.~\ref{fig:2} (b), $\mathcal{A}_{\text{exc}}$ achieved an 84.8\% approval rate (39/46), correctly modeling pathological relationships such as mutual exclusivity (e.g., hemorrhage vs. infarction), coexistence, and causality (e.g., atrial fibrillation leading to stroke). The remaining 15.2$\%$ of errors were primarily attributable to oversimplified modeling of complex complications.
$\mathcal{A}_{\text{conf}}$ demonstrated exceptional utility with a 97.6\% (362/371) success rate in identifying critical differential pairs, such as distinguishing BPPV from Meniere’s disease based on positional triggers. Only 2.4\% were mislabeled. Furthermore, $\mathcal{A}_{\text{adj}}$ achieved 74.2\% (144/194) accuracy in executing logical weight modifications. These results confirm that RE-MCDF's gains stem from rigorous clinical reasoning rather than simple probabilistic pattern matching.

\subsubsection{Qualitative analysis of cases}
%Fig.~\ref{fig:example} illustrates the diagnostic trajectory of RE-MCDF for a 78-year-old female patient with acute vertigo and a critical CT finding of cerebellar hemorrhage (1.9 cm $\times$ 1.5 cm). RE-MCDF demonstrates three key advantages: First, $\mathcal{A}_{\text{lab}}$ correctly prioritizes ``cerebellar hemorrhage'' as the critical abnormal entity (weight 0.55). This mechanism guides the evidence evaluation toward high-value imaging findings, preventing diagnostic drift caused by non-specific symptoms. Second, $\mathcal{A}_{\text{exc}}$ identifies the \textbf{pathological exclusivity} between cerebral hemorrhage and infarction. Consequently, prompted $\mathcal{A}_{\text{adj}}$ to provide a pathological logic score of 0.0, effectively eliminating misdiagnosis risks based on shared symptoms (e.g., nausea). Third, the KG supplement module ($k_{\text{sup}}=1$) successfully introduces ``hypertension'' as a candidate diagnosis, which ranks $4^{th}$ (score 0.407). This aligns with the ground truth of ``Hypertension Grade 3,'' demonstrating the value of KG in comorbidity identification. The final ranking (Cerebral Hemorrhage 0.920 $>$ Hypertensive Encephalopathy 0.839 $>$ Cerebellar Dysfunction 0.672 $>$ Hypertension 0.407) is highly consistent with the ground truth. The Top-1 diagnosis precisely matches the ICD-10 code for cerebellar hemorrhage, reflecting a logical clinical hierarchy: the acute hemorrhage event as the primary cause, with hypertension as the underlying condition.
Fig.~\ref{fig:example} illustrates RE-MCDF's diagnostic trajectory for a 78-year-old female with acute vertigo and CT-confirmed cerebellar hemorrhage. This case highlights three key strengths of the framework: \textit{(i)} \textit{Priority Weighting:} $\mathcal{A}_{\text{lab}}$ assigns a high weight (0.55) to the critical indicator \textit{cerebellar hemorrhage,} preventing diagnostic drift toward nonspecific symptoms such as nausea or dizziness; \textit{(ii)} \textit{Logic Enforcement:} $\mathcal{A}_{\text{exc}}$ identifies the \textit{mutual exclusivity} between hemorrhage and infarction; accordingly, $\mathcal{A}_{\text{adj}}$ assigns a 0.0 compatibility score, effectively mitigating symptom-driven misdiagnoses (e.g., nausea); \textit{(iii)} \textit{MKG Augmentation:} The supplement module identifies hypertension ($4^\text{th}$, score 0.407), matching the ground truth \textit{hypertension grade 3.} The final ranking (cerebral hemorrhage 0.920 $>$ hypertensive encephalopathy 0.839 $>$ hypertension 0.407) closely aligns with the clinical hierarchy, with the acute event as the primary diagnosis and hypertension as the etiological factor.

\section{Conclusion and Future Work}
In this paper, we presented RE-MCDF, a relation-enhanced multi-expert framework that addresses evidence heterogeneity and logical inconsistencies in EMR reasoning. By introducing a \textit{generate-verify-revise closed-loop} mechanism, RE-MCDF emulates the deliberative and collaborative workflow of real-world clinical consultations.  Experimental results on the NEEMRs and XMEMRs datasets demonstrate that RE-MCDF outperforms the existing baselines. Three avenues of research constitute promising future directions: 
\textit{(i)} Integrating medical imaging (e.g., CT/MRI) with textual EMRs to enhance diagnostic depth; \textit{(ii) }Validating the framework's efficacy across other complex medical specialties beyond neurology; \textit{(iii) }Optimizing reasoning efficiency for real-time integration into hospital decision support systems.

\bibliographystyle{IEEEtran}
\bibliography{refs}
\end{document}